# A proposition of a robust system for historical document images indexation

Nizar ZAGHDEN[1,2]
1 REGIM: Research Group on Intelligent Machines,
University of Sfax, ENIS,
Department of Electrical Engineering.
BP W - 3038, Sfax, Tunisia

nizar.zaghden@ieee.org

, Remy MULLOT[2],
2 L3I: Laboratoire Informatique1 Image Interaction
Université de La Rochelle,
France, BP 17042

remy.mullot@univ-lr.fr

Adel M.ALIMI
1 REGIM: Research Group on Intelligent Machines,
University of Sfax, ENIS
Department of Electrical Engineering
BP W - 3038, Sfax, Tunisia

adel.alimi@ieee.org

**Abstract**

*Characterizing noisy or ancient documents is a challenging problem up to now. Many techniques have been done in order to effectuate feature extraction and image indexation for such documents. Global approaches are in general less robust and exact than local approaches. That's why, we propose in this paper, a hybrid system based on global approach (fractal dimension), and a local one, based on SIFT descriptor. The Scale Invariant Feature Transform seems to do well with our application since it is rotation invariant and relatively robust to changing illumination. In the first step the calculation of fractal dimension is applied to images, in order to eliminate images which have distant features than image request characteristics. Next, the SIFT is applied to show which images match well the request. However, the average matching time using the hybrid approach is better than "fractal dimension" and "SIFT descriptor" techniques, if they are used alone.*

## 1. Introduction

Nowadays a lot of information is still stored in libraries and great effort must be done to digitalize or extract features from the huge quantities of old documents. When talking about images containing mostly textual information, OCR systems can be applied to characterize image documents. But these Character Recognition Systems seems to fail when document images are ancients or even noisy. Many researches have been done to characterize old documents in different origins (latin, arabic, chineese…). The recognition of different classes in historical documents requires suitable techniques in order to identify similar classes. As contemporary documents, techniques dealing with global features can be applied to heterogeneous type of documents. But extracting local features from images differs from the language of the text written in documents. So, the application of methods based on local features may fail when it is applied to heterogeneous types of documents. We propose in this paper a new method based on both, global and local features (figure 1).

This paper is organized as follows. Section 2 presents our image indexation approach in details. Section 3 reports the experimental results. Section 4 concludes the paper.

## 2. Our Image indexation approach

We first introduce the phases which we followed in our approach. In fact, we have segmented manually about 1000 images issued from the CESR base with a resolution of 300 dpi each. The specificity of this base is that it is heterogeneous, and contains figures, different fonts. It deals with ancient documents, and as we know almost of techniques which suppose to have good results in contemporary documents may fail within this application. In this work, we are interested in textual content, and we apply the fractal dimension as a global approach in the first step.

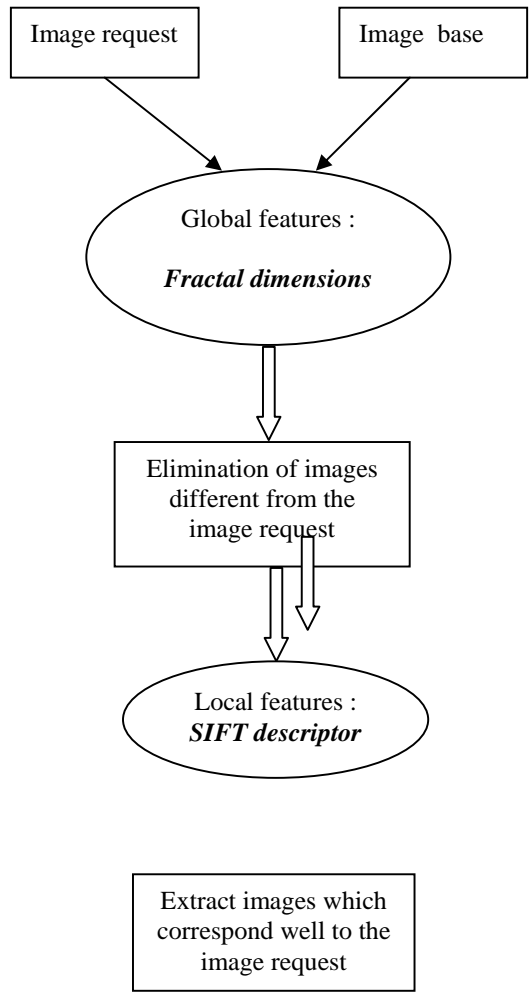

Figure 1: global scheme of the proposed method

The global features of an image are often used by many researchers in the image retrieval domain. The global approach cannot represent image details or regions, particularly robustness to partial visibility and high informational content. For this, we propose in this work an hybrid approach combining both global and local features [6].

## 2.1 Fractal dimension

The fractal dimension is a useful method to quantify the complexity of feature details present in an image. Until today there is no common definition of what is fractal, but it is clear that fractal has many differences with Euclidean shapes. The fractal dimension is the main characteristics of fractals and it is assumed that it exceeds strictly topological dimension of fractal sets. In this paper we propose a new algorithm to estimate the fractal dimension of images and we compare this method with existing methods.

There are mainly two different methods to calculate fractal dimension: Box Counting and Dilation methods [9]. Several algorithms are been derived from the box counting approach such as differential box counting [7] and the reticular cell counting [1]. The main idea in Box counting algorithms is to divide images by similar box sizes. Then the fractal dimension of the set can be estimated by the equation:

D=log (Nr)/log (1/r), Where Nr represents the number of boxes comprising the sets each scaled down by a ratio r from the whole. Sarkar and Chaudhuri [7] proposed the differential box counting approach (DBC), which add a third coordinate for 2D images, corresponding to the gray level value of boxes. In each box (i, j), the authors calculate the maximum and minimum gray values: L and K. Then the value of gray value to be considered for that box is:

$n_r(i, j) = l-k+1$.

The total contribution of gray value of the image is the sum of nr (i, j). The new method that we propose for estimating fractal dimension is derived from the latter method and it is called the CDB method (Comptage de Densité par Boîte). We consider that the image of size M×M pixels has been scaled down to a size s×s where M/2 >s>1 and s is an integer.

Then we have an estimation of r = s / M. The (x, y) space is partitioned into boxes (i, j) of size s×s. On each box we calculate the density of black pixels $n_r(i, j)$.

$Nr = \sum_{i,j} n_r(i, j)$ represents the total contribution

of the image.

We have calculated fractal dimensions of images by three different methods: dilation method, Differential Box Counting (DBC) [8] and our own approach (CDB: Comptage de densité par Boîte). We have chosen to calculate fractal dimension for the DBC and CDB method by using five different parameters corresponding to the maximum size box: 10, 15, 20, 30 and 40. For the dilation method we have calculated fractal dimension of images by using five values of order dilation: 5, 10, 15, 20, and 30.

Many studies have been taken to define the limit of box sizes [1, 3, 7]. Referring to [1] the maximum box size is: L=M/G, where M represents image size and G corresponds to the number of different gray levels in the image. Sarkar et al [7] have considered M/2 as the maximum size of boxes for an M×M image size. In our approach (CDB) the images are converted to binary images. Then the number of gray levels is two. The maximum box size is M/2. It is clear that this limit verifies well the size limit of boxes proposed by both Bisoi and Sarkar.

Concerning the dilation method we have chosen to consider low values of the order dilation. In fact the most of researchers who calculate fractal dimension use box counting method in their approach and there is not a big interest on the dilation method. In previous work [8], we proved the importance of results obtained by our contribution in calculating fractal dimension using the CDB method in comparison with similar methods applied for the same standard images issued from Brodatz images [9].

The fractal dimensions values calculated for every image in the base is used here as a first step of the indexation process. In fact, the CDB applied here is considered as a global approach since it gives indices of the whole image [2].

As shown in figure 2, the fractal dimension is used to classify images and to reduce the number of images on which we will compare their local features. For every image request we calculate her fractal indices, and then we estimate the images which have fractal dimensions close to the first one. We fixed a manually threshold=0.5, in order to reject images having an absolute difference of fractal dimension than the image request over than the threshold. The choice of this method is justified by the fact of it's robustness for characterising old documents [2] and also for the fast execution time needed to manipulate a number of 1000 images in our base.

In the figure 3, we consider only the group of images 1 and 2, but we reject the group image 3, which have fractal features distant from the image request features.

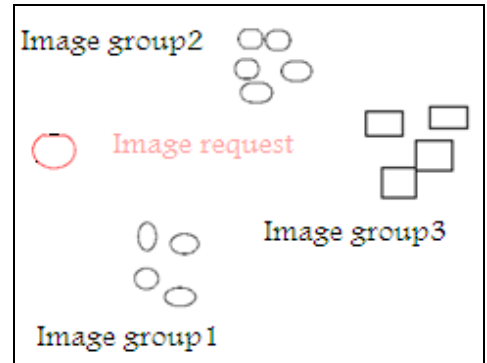

Figure 2: classification of images with fractal dimensions

In the second step, we consider just the image group 1 and the image group 2. The use of interest points in image documents matching allows us to use local properties of the image. We apply for these images the SIFT descriptor in order to obtain the ordering images which match the image request. This step is so helpful for us since we deal with a big number of documents. The number of remaining images is not usually constant because it depends on the image request.

### 2.2 SIFT Descriptor

The SIFT descriptor is based on the gradient distribution in salient region, and constructed from 3 D histogram of gradient locations and orientations [4]. Sift feature is invariant for rotation, scale changes, and illumination changes. A 128 dimension vector representing the bins of the oriented gradient histogram is used as descriptor of salient feature [11]. When dealing with a huge database, SIFT descriptor is a significant drawback. In the method proposed in this paper, the high dimensionality of the base is reduced in the first step by eliminating far images. The SIFT produces several features, yielding to a large feature space, which needs to be searched, indexed and matched. The interest points produced by SIFT are more dependent on structure than on illumination. They are widely used in several computer vision and pattern recognition tasks. However, as the number of interest points per image varies from a minimum of 100 keypoints to over than 3000 keypoints for our images of 512 * 512 pixels. This difference in the number of generated keypoints is that images which we treat have different contents as shown in figure 3.

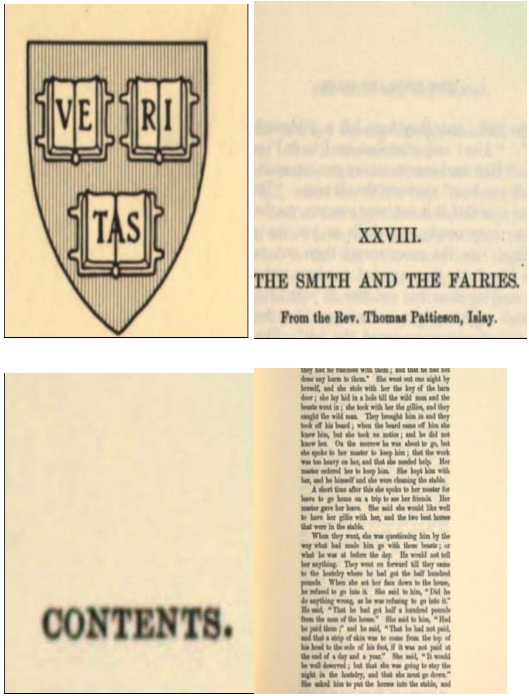

Figure 3: samples of historical document images

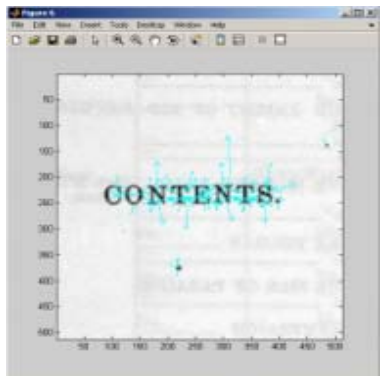

Figure 4: keypoints extracted of an old image document

The SIFT points are local extremes (minimum and maximum) in a scale space [11] composed by differences of Gaussians of progressively larger standard deviations, as shown in figure 4.

The choice of SIFT descriptor consists in the fact of using local features in order to characterize well images [8].

The application of SIFT leads to the number of matched points between two images, but we can also make some ameliorations to indicate the best images which suits the image request, as we will explain in the next section.

## 3. Experimental results

Image indexation is a fundamental technique in many applications of computer visions such as classification, object recognition, 3D reconstruction [9]. The main idea of the matching process is to identify if two images may correspond, based on the comparison of each features extracted from the two images. For correct word matching, the conventional analysis of distances between text objects needs very much time of calculation especially in our case where we have a huge old document base. This base contains documents issued from different centuries and having various characteristics such as language, noise degree, scripts and figures.

Intensity based techniques may fail in almost of cases because they are sensitive to scaling variations and illumination changes. But the local invariant descriptors, such as the SIFT used in this paper, are very robust to the possible variations and transformations in images [11].

We applied the SIFT descriptor after the reject phase obtained by the description of fractal dimensions calculated for all images in our base. This collection of images contains about 1000 images of 512*512 pixels each. But the difference between images is their heterogeneity, and we can find only one word in an image, where in other cases we can find a lot of words or figures in the same image (figure 3).

The SIFT extract the number of matching points between two images. A point is considered as matching point only if the nearest neighbor of the feature of the image 1 corresponds to the features of image B, and also the distance between them is under a threshold. In the SIFT descriptor the nearest neighbor is calculated by the Euclidean distance [4].

We calculate the matching points between the image request and each of the images remaining after the first phase.

We used the term of similarity introduced by Maatar et al in [5]. The image similarity can be defined as a mean value between the number of matched m points of (image 1, image 2) and those from (image 2, image 1), since the number of matched points differs in the two cases.

$$M(I1, I2) = \frac{m(I1, I2) + m(I2, I1)}{2},$$

Where M(im1, im2) corresponds to the number of matches between im1 and im 2 as presented in figure 5.

A new similarity measure can be added in order to assume the class to which the input image can belong. The measure of similarity between an image I and a class A that contains n images Ai as:

$$S(I, A) = \sum_{i=1}^{n} M(I, A_i).$$

In figure 2 we presented the images belonging to three different classes. The image group 3 is rejected by the first phase of global features. To assume that the input image belongs to image group 1 or image group 2, we calculate the term S. If this term indicates us that S(I, image group 1) > S( I, image group 2), then we can assume that the input image belongs to the first class. We can also, reject all the images belonging to the second class.

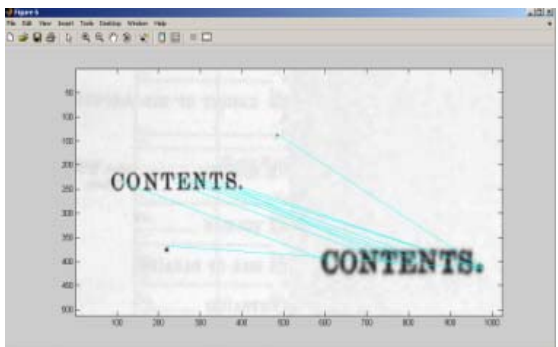

(a)

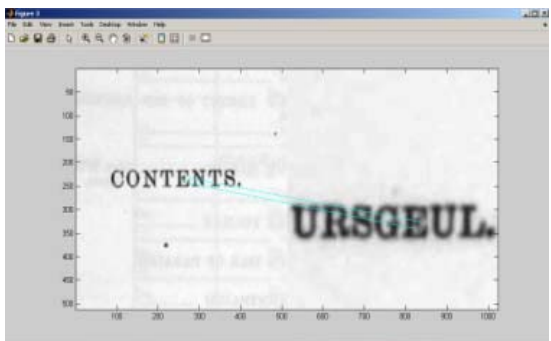

(b)

Figure 5: matching points; a: 16 matching points, b: 2 matching points

In the table below, we present the result of matching points for the two images input containing respectively the words: "contents" and "URSGEUL".

Table 1: Number of matching points for two images request

|  | Corresponding Image 1 | Corresponding Image 2 |
|---|---|---|
| Image 1 | 12 | 2 |
| Image 2 | 16 | 10 |

As we can see, the best number of matching points for the image 1 is 12, and 16 for the second image. The matching points of this image are shown in figure 6.

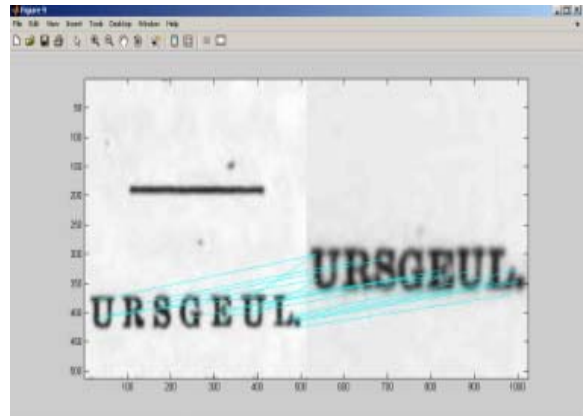

Figure 6: the presence of 16 matching points for the image2

We can assume here, that the number of prototypes that correspond to the image requests is not very high, but this number would be highest if we choose other images from other collections of ancient documents such as " Madonne library" or " British library", which may have more similarities with images in our base.

It's obvious to notice that the best number of matching points can't affirm that two images contain the almost similar words (figure 7).In this figure, we obtained for the image (a), a 19 matching keypoints and a 90 similar keypoints in the image (b). In fact, with keypoints descriptors, we compare features pixel by pixel and not for a whole word or phrase.

As second experimentations, we propose to compare a simple word image with images containing many words or paragraphs in order to compare the matching points resulting, or if the input image may exist in the second image as illustrated in figure 8. We can see in the figure 8 that SIFT descriptor performs well the matching of pseudo-word image containing "en" in the second image.

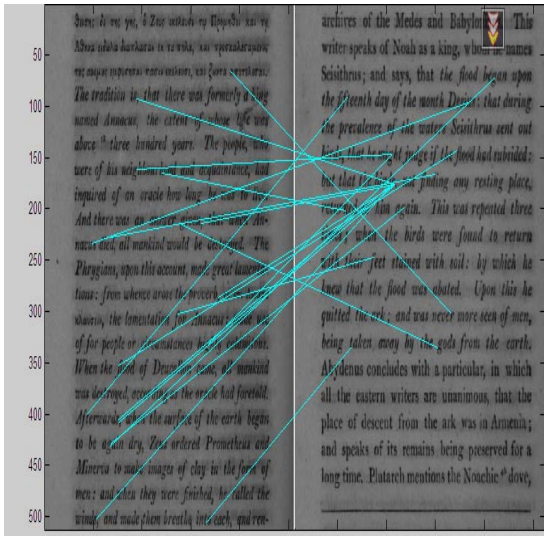

(a)

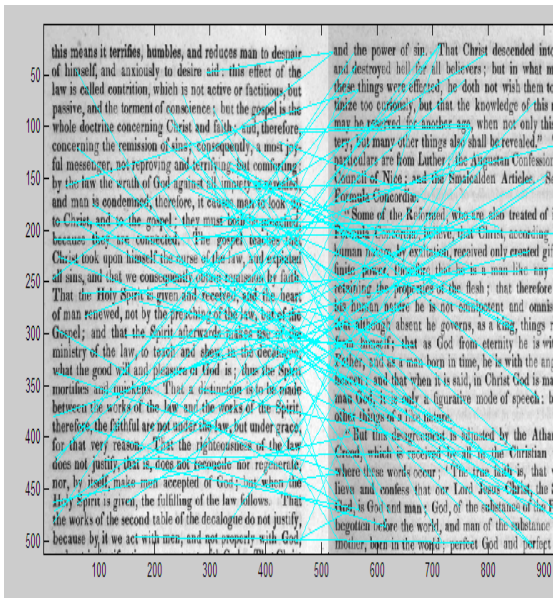

(b)

Figure 7: Matching points in whole documents

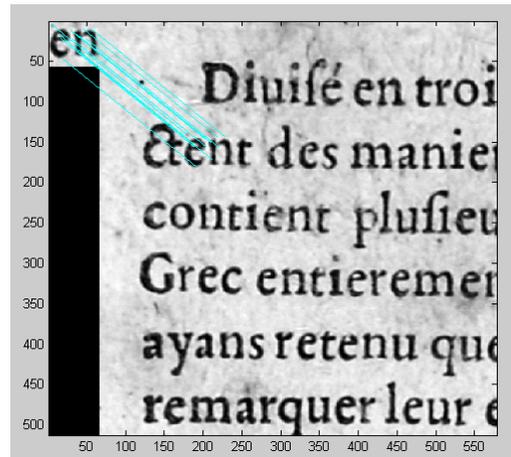

(a)

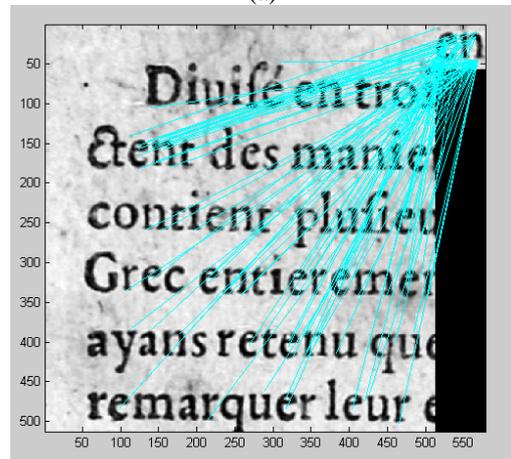

(b)

Figure 8: retrieval of the image of pseudo-word "en" in the second image document

In the above figure, we can notice a clear difference between the matching points calculated in figure (a) and in figure (b). In fact the value of matching points is 13 in the first and 99 in the second test. This difference explain the fact of using the mean of matching points in the two tests as explained by Maatar et al in [5].

In this paper we treated essentially old document text images, but we applied also our system for figures of old documents. We can assume from the figure 9 that the results obtained for such types of ancient documents are also as important as the ancient text images as illustrated in previous works [10]. The number of final matching points M is about 32 since m (I1, I2) =31 and m (I2, I1) =32.

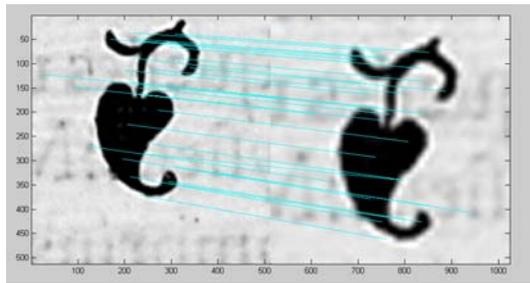
(a)

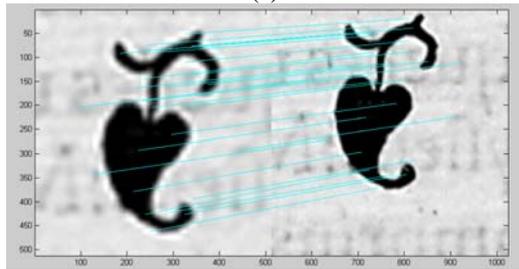
(b)

Figure 9: Matching points in old documents figures

## 4. Conclusion

We presented in this paper a robust method for indexing ancient documents. This hybrid method combines local and global image features. The global approach is used as first step to classify images and to reject distant images from the image request. The use of SIFT descriptor for the second step of our methods allows us to obtain the best corresponding images to the input image. The main advantage of using SIFT descriptor as second phase is the gained time than matching one image with 1000 other images that we have in our base. Combining fractal dimensions with keypoints descriptor is helpful to produce more effective and more efficient solution to the issue of image matching. Several tests were applied and they proved the robustness of this system such as for text images or figures of ancient documents.

## Acknowledgements
All the authors of this paper are grateful and thank the committee members of Piranna Project CMCU, 08G1416.